\pdfoutput=1

\documentclass[11pt]{article}

\usepackage[]{acl}

\usepackage{times}
\usepackage{latexsym}

\usepackage[T1]{fontenc}

\usepackage[utf8]{inputenc}

\usepackage{microtype}

\usepackage{graphicx}
\usepackage{booktabs}
\usepackage{array}
\newcolumntype{L}[1]{>{\raggedright\let\newline\\\arraybackslash\hspace{0pt}}m{#1}}
\newcolumntype{C}[1]{>{\centering\let\newline\\\arraybackslash\hspace{0pt}}m{#1}}
\newcolumntype{R}[1]{>{\raggedleft\let\newline\\\arraybackslash\hspace{0pt}}m{#1}}
\usepackage{multirow,tabularx}
\usepackage{amsmath}
\usepackage{amssymb}
\usepackage{bbm}
\usepackage{xcolor}

\title{Informative Text Generation from Knowledge Triples}

\author{Zihao Fu$^\dagger$, Yijiang River Dong$^\clubsuit$, Lidong Bing$^\ddagger$, Wai Lam$^\spadesuit$ \\
  $^\dagger$Language Technology Lab, TAL, University of Cambridge, 
  $^\clubsuit$University of Pennsylvania\\
  $^\ddagger$DAMO Academy, Alibaba Group, 
  $^\spadesuit$The Chinese University of Hong Kong \\
  \texttt{zf268@cam.ac.uk, riverd@sas.upenn.edu}\\
  \texttt{l.bing@alibaba-inc.com, wlam@se.cuhk.edu.hk} }

\begin{document}
\maketitle
\begin{abstract}
As the development of the encoder-decoder architecture, researchers are able to study the text generation tasks with broader types of data. Among them, KB-to-text aims at converting a set of knowledge triples into human readable sentences. In the original setting, the task assumes that the input triples and the text are exactly aligned in the perspective of the embodied knowledge/information. In this paper, we extend this setting and explore how to facilitate the trained model to generate more informative text, namely, containing more information about the triple entities but not conveyed by the input triples. To solve this problem, we propose a novel memory augmented generator that employs a memory network to memorize the useful knowledge learned during the training and utilizes such information together with the input triples to generate text in the operational or testing phase. We derive a dataset from WebNLG for our new setting and conduct extensive experiments to investigate the effectiveness of our model as well as uncover the intrinsic characteristics of the setting.
\end{abstract}

\section{Introduction}
Text generation is a longstanding NLP task, and congenitally it focuses on addressing two questions separately: what to say and how to say \cite{Reiter_building_1997,Jurafsky:2000:SLP:555733}. Accordingly, earlier systems usually have three components: content selection, macro/micro-planning, and surface realization. The recent development of the encoder-decoder architecture allows researchers to perform text generation from richer input data. The KB-to-text generation problem, e.g. the WebNLG task \cite{gardent2017creating, gardent2017webnlg}, aims at directly converting a set of KB triples into human-readable sentences. For example, given the knowledge triples ( $<$ Bill Gates, Birthplace, Seattle $>$, $<$ Bill Gates, FounderOf, Microsoft $>$), the goal is to generate a comprehensible sentence such as ``\textit{Bill Gates, the founder of Microsoft, was born in Seattle.}''

\begin{figure}[!t]
\centering
\includegraphics[width=0.8\columnwidth]{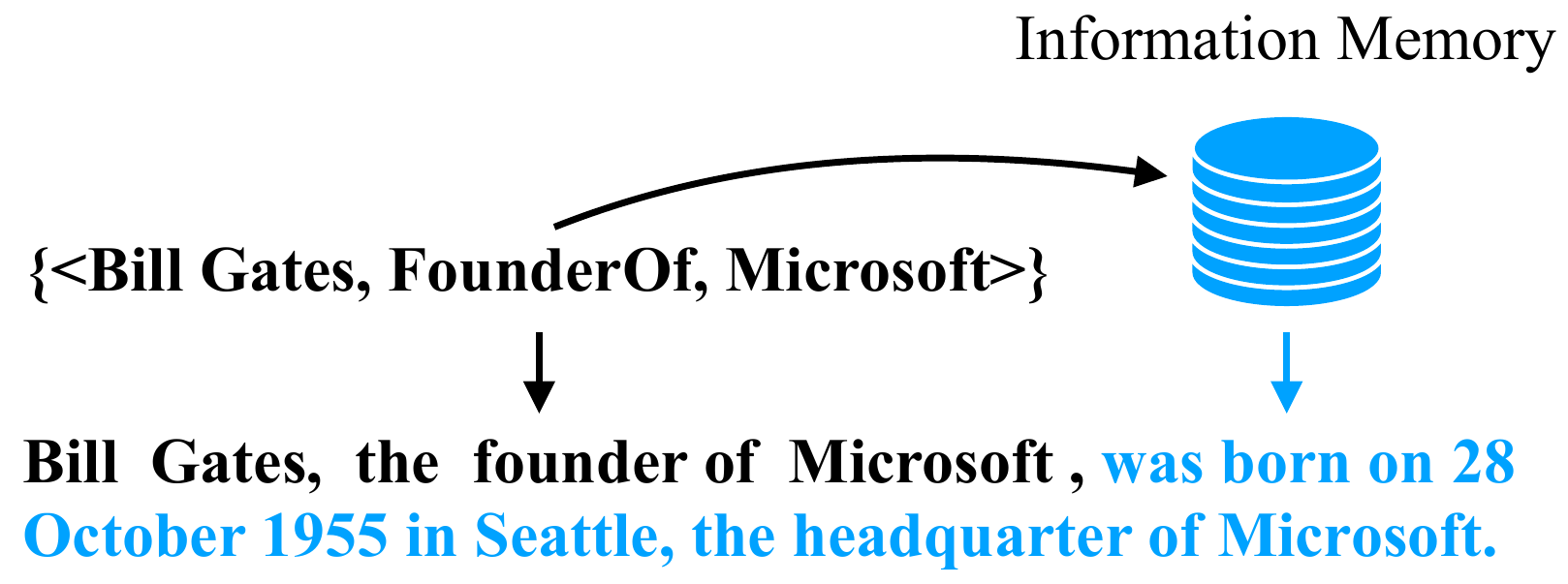}
\caption{The proposed task: generating text based on the input triples together with the information memory.}
\label{fig:problem}
\end{figure}

In the current KB-to-text setting, it assumes that the input data (i.e. triples) and the output text are exactly aligned. In other words, the generated text only contains the information conveyed by the input triples. However, this setting is somewhat rigid and less practical. When human beings interpret the data via natural language to others, they are prone to adding extra information according to some common experience to make the sentence more comprehensive. For example, when humans are explaining the triple $<$Bill Gates, FounderOf, Microsoft$>$, they may add extra words to tell that the headquarter of Microsoft is exactly the place where Gates was born. Thus the whole text is more informative especially for those readers who know less about Bill Gates or Microsoft. Therefore, assuming that the data and the text are exactly aligned mismatches with how humans usually write sentences.

In this paper, we propose a new problem named Informative Text Generation (ITG). It extends the current KB-to-text task to a setting that requires to generate text from an input triple set together with a information repository. It should generate more informative text containing more information about the triple entities that is not conveyed by the input triples.
As shown in Fig. \ref{fig:problem}, when the input triple is ``$<$Bill Gates, FounderOf, Microsoft$>$'', the ITG task is to generate a sentence as exemplified by ``\textit{Bill Gates, the founder of Microsoft, was born on 28 October 1955 in Seattle, the headquarter of Microsoft.}'' Note that though the generated text may not be exactly the same as the example text, it should be a sentence containing more information about ``\textit{Bill Gates}'' and ``\textit{Microsoft}'' in addition to the fact that Bill Gates is a founder of Microsoft.
Our new setting requires the trained model to retrieve relevant knowledge automatically from the memory and generate text containing some extra narrative that is not conveyed in the input triples. Therefore, the generated sentences are more informative. Obviously, existing data-to-text models cannot handle our new problem.

In order to solve the ITG problem, we propose a novel information Memory Augmented Generator (iMAG). During the training, the model randomly drops some input triples and store the information that is conveyed in the text but not corresponding to any remaining triples as a background knowledge into the information memory. It also learns to retrieve this kind of knowledge automatically from the memory. Afterwards, the model will be able to generate text with the presence of the memory providing extra knowledge about the input entities in the operational or testing stage. 
Specifically, our model is composed of two components: the information memory and the sequence-to-sequence (seq2seq) network. 
Within the information memory component, part of the input triples is dropped randomly in the training phase making the target sentence containing more information than the remaining input triples. Then, we design a querier that can fetch the related information from the memory according to the remaining input triples. The seq2seq network generates output sentences by considering both the fetched information and the remaining input triples. 
To avoid the tendency that the fetched information may repeat some information conveyed by the input triples, we propose several ways to suppress the repetitions. 
We derive a dataset from the WebNLG dataset suitable for conducting experiments for our problem setting  . If needed, other datasets can be derived in a similar way from a source dataset with similar nature. We compare our proposed model with some existing models adapted to this new setting. The experimental results show that our model has a better capability of capturing the extra information encountered in the training data and thus generates more informative text.

\section{Related Works}
Recently, a few data-to-text tasks have been proposed which aim to generate text from some formatted data such as tables and knowledge triples. 
The WebNLG task \cite{gardent2017creating, gardent2017webnlg,tran2020webnlg} generates sentences corresponding to a set of related triples sampled from DBpedia \cite{auer2007dbpedia, lehmann2015dbpedia}. \citet{lebret2016neural} propose to generate people's short biography from Wikipedia infobox. \citet{novikova2017e2e} propose to generate restaurants's reviews based on corresponding attributes.
\citet{wiseman2017challenges} and \citet{chen2008learning} generate match summaries for ball games, while \citet{liang2009learning} generate  weather forecast.
Our task in this paper is an extension of this setting, particularly WebNLG (other dataset can be adopted easily in a similar manner). The difference is that these existing settings do not generate text containing extra knowledge that is not mentioned in the input data.

The memory mechanism \cite{weston2014memory,graves2014neural,sukhbaatar2015end} have been proposed in many Neural Machine Translation (NMT) systems. \citet{wang2016memory,meng2018neural} propose to use memory to enhance the decoder in the seq2seq framework. \citet{miller2016key} propose key-value memory networks to read and comprehend documents in a QA framework while \citet{chen2019enhancing,yang2019enhancing} propose to utilize specific external knowledge triples and ConceptNet explicitly to help generate which is restricted to specific areas. \citet{shen2019pragmatically} propose to use a pragmatic method to ensure that the information is fully conveyed. We propose to utilize an embedded memory to add more relevant descriptions.

It has been observed in many works
\cite{mi2016coverage,tu2016modeling,wu2016google,holtzman2019curious} that many generation systems suffer from the content-repeating problem. Actually, this problem occurs because the attention focuses on the same region many times. \citet{mi2016coverage} and \citet{tu2016modeling} propose the coverage mechanism to handle this problem while \citet{meng2018neural} propose to use memory-based attention to eliminate the content-repeating problem. 
However, in our problem, the content-repeating problem is mainly caused by the information overlapping between the input source and the information memory. Therefore, simply averaging the attention cannot handle our problem.

\section{The Proposed Model}

We use $h$, $r$ and $t$ to denote the head entity, the relation and the tail entity respectively in a knowledge triple. Given an input triple set with $c$ triples $G=\{<h_1,r_1,t_1>,\cdots,<h_c,r_c,t_c>\}$, our information Memory Augmented Generator (iMAG) aims at generating a text sequence $S=(s_1,s_2,\cdots,s_m)$ that maximizes the conditional probability given the input triple set $G$ and the information memory $M$:
\begin{equation}
    \max_{\theta,\psi} p_{\theta} (S|G,M_\psi),
\end{equation}
where $\theta$ represents all learnable parameters of the network and $\psi$ stands for the memory parameters. 

\begin{figure}[!t]
\centering
\includegraphics[width=0.8\columnwidth]{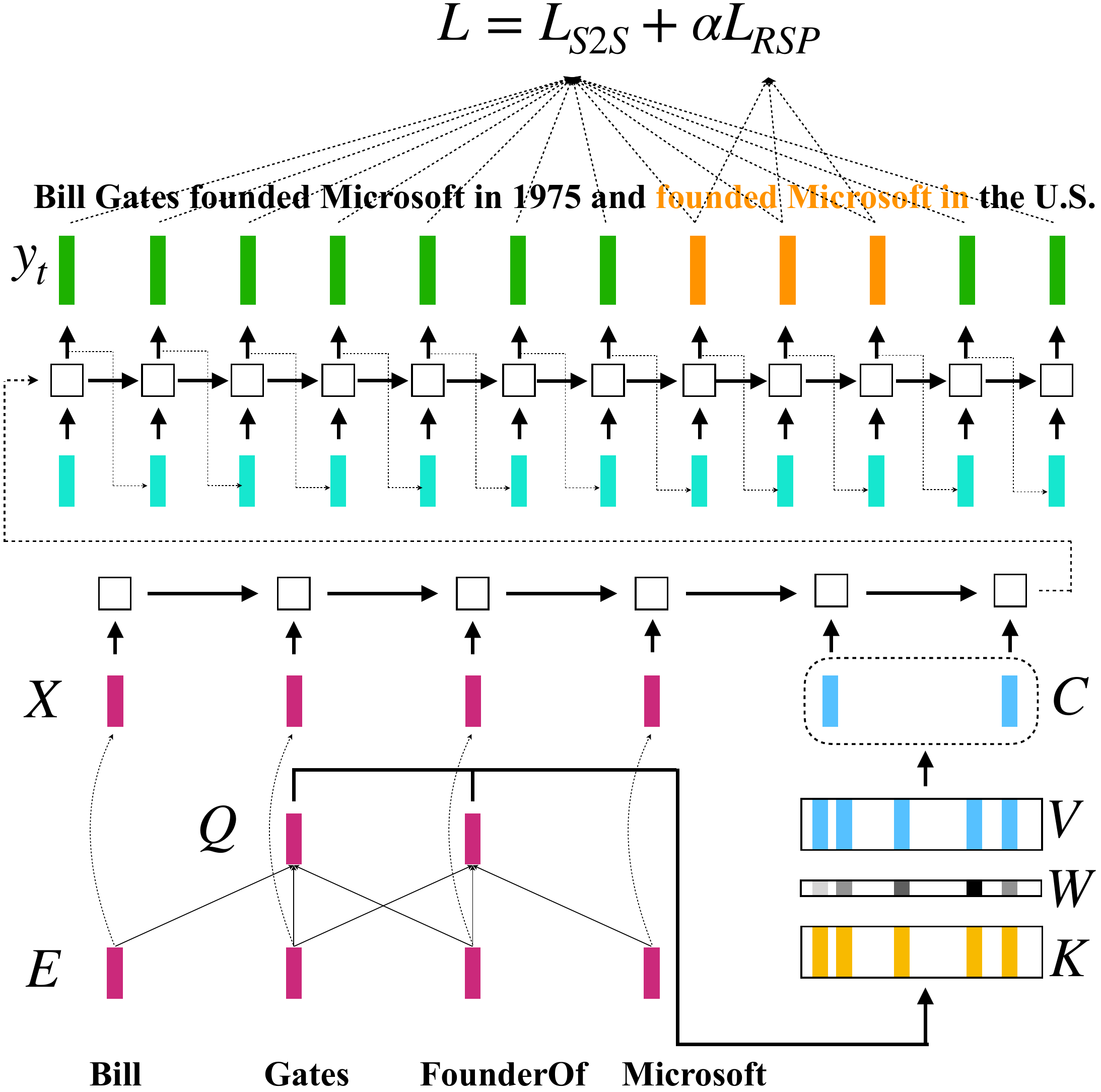}
\caption{Framework of our model. We omit some components such as attention and copy for clarity.}
\label{fig:framework}
\end{figure}

As illustrated in Fig. \ref{fig:framework}, our model contains two components.
In the \textbf{Information Memory} component, a memory matrix $V$ with each column $V_i$ as a memory slot is designed to memorize the extra information during the training. Given an input triple set $G$, the model firstly removes part of the input triples by a remaining ratio. Then, it generates a querier matrix $Q$ by sliding a CNN window over the flattened input triples to get the querying vector $Q_j$ for the $j$th window. We get the relevant score  $W_j$ of each memory slot for the $j$th window and get the relevant knowledge vector $C_j$ by a weighted sum of the memory slots.
The relevant knowledge matrix $C$ composed of all $C_j$ as its columns is then concatenated to the embedding of the original input sequence and fed into the subsequent component. A
\textbf{sequence-to-sequence network} first encodes the sequence into hidden vectors and then its decoder generates the output text. The attention \cite{bahdanau2014neural} and copy \cite{gu2016incorporating} mechanisms are used to enhance the performance.
Since the information memory may return hidden representations containing the same information with the input triples and resulting in the content-repeating problem, we design several variants of iMAG to alleviate the problem, such as the repeating sentence penalty (i.e. $\mathcal{L}_{rsp}$ at the top of Fig. \ref{fig:framework})

\subsection{Information Memory}
Given an input triple set, the information memory firstly removes some triples according to a preserving ratio $\zeta$ in the training phase (it does not remove anything when testing). Then, it flattens the remaining triples as a word sequence:  $G'=[h_1^{(1)},\cdots,$ $h_1^{(n_{h_1})},r_1^{(1)},\cdots,r_1^{(n_{r_1})},t_1^{(1)},\cdots,t_1^{(n_{t_1})},\cdots,h_{c'}^{(1)},\cdots,$ $h_{c'}^{(n_{h_{c'}})},r_{c'}^{(1)},\cdots,r_{c'}^{(n_{r_{c'}})},t_{c'}^{(1)},\cdots,t_{c'}^{(n_{t_{c'}})}]$, where $h_i^{(j)}$ stands for the $j$th word of the head entity in the $i$th remaining triple in $G$. $n_{h_i}$ stands for the number of words in the $i$th remaining entity $h_i$ and $c'$ is the number of the remaining triples. Hereafter, we denote $G'=[x_1,x_2,\cdots,x_n]$ for clarity in which $x_i$ stands for each word in $G'$ and $n$ is the word count in $G'$. Then $G'$ is mapped into vectors by an embedding layer, $E=Emb(G')$,
where $E\in \mathbb{R}^{e\times n}$ while $e$ is the embedding dimension.

We extract the feature of each segment of the sequence by applying a CNN module on the embedding $E$, namely, $Q = CNN(E)$.
The kernel size of CNN is set to $e\times k$ while $k$ is the kernel length. Therefore, the CNN module encodes each window of $k$ input words into a hidden representation and use them to retrieve a memory vector related to those words. The $j$th column of the matrix $Q\in \mathbb{R}^{e\times(n-k+1)}$ represents the query vector of the $j$th window. $Q$ will be used as a querier to get the relatedness of each memory slot by a matrix product as $\tilde{W}=K^TQ$,
where $K\in\mathbb{R}^{e\times l}$ is a learnable matrix used to calculate the relatedness of each memory slot and $l$ is the total memory slot number which is design to be a tunnable hyperparameter. $\tilde{W}\in \mathbb{R}^{l\times(n-k+1)}$ is the relatedness of $n-k+1$ windows to the $l$ memory slot, it will be further normalized with a softmax layer as:
\begin{equation}
    W_{ij}=\frac{\exp{(\tilde{W}_{ij}})}{\sum_{i=1}^l \exp{(\tilde{W}_{ij})}},
\end{equation}
in which $W\in \mathbb{R}^{l\times(n-k+1)}$ has the same size as $\tilde{W}$. The only difference is that each column of $W$ sums up to $1$ and can be therefore regarded as the probability distribution of each memory slot.

With the probability of each memory slot, the relevant knowledge matrix  can be calculated by a weighted sum as $C=V^TW$,
in which $V\in\mathbb{R}^{l\times e}$ is the memory matrix. $C\in\mathbb{R}^{e\times (n-k+1)}$ is the weighted sum of the memory values while the $j$th column $C_i$ is the relevant knowledge for the $j$th sliding window.
The output of the information memory is a simple concatenation of the original input embedding $E$ and the memory information $C$, namely, $X=[E;C]$,
in which $X\in\mathbb{R}^{e\times (2n-k+1)}$ will be fed into the subsequent seq2seq network. It contains the information of both the input triples and the fetched memory and thus it can generate sentences with more information than merely the input triples.

\subsection{Sequence-to-Sequence Network}
The seq2seq takes $X$ as input. We use attention to enhance the network's expression capability. 
In the encoder, $X$ is fed into an LSTM layer to get the hidden representation: $H=LSTM(X)$,
in which $H\in\mathbb{R}^{e\times (2n-k+1)}$ has the same size with $X$. We denote $h_{-1}$ as the last hidden vector of $H$. $h_{-1}$ aggregates the information from the entire input sequence and thus can be regarded as the context vector. 

In the decoder, the hidden state $u_t=LSTM([u_{t-1};z_{t-1}])$ is calculated based on the last output $z_{t-1}\in \mathbb{R}^{e}$ and the last hidden state $u_{t-1}$, where $u_1=h_{-1}$. $u_t$ is the hidden representation of the generated sequence, and it will be refined by an attention layer together with the input hidden representation $H$. The calculation is denoted as $z_t=Attn(u_t,H)$,
where $Attn$ is the attention function \cite{bahdanau2014neural} which considers $u_t$ and its relation to each $H_i$ to calculate $z_t$. The word probability distribution $y_t$ is calculated as $y_t=\text{softmax}(M_pz_t+b_p)$, in which $y_t\in \mathbb{R}^{|V|}$  and $|V|$ is the vocabulary size. $M_p\in \mathbb{R}^{|V|\times e}$ is a transformation matrix and $b_p\in\mathbb{R}^{|V|}$ is the bias vector. 
In the training, our model minimizes the following negative log likelihood loss:
\begin{equation}
    \mathcal{L}_{s2s}=-\sum_{t=1}^T\log y_t[s_t],
\end{equation}
in which $s_t\in S$ is the $t$th gold standard target word and $y_t[s_t]$ denotes the $s_t$th element of $y_t$. We also apply the copy mechanism \cite{he2017generating} to enhance the  model capability.

\subsection{Variants for Handling Content-Repeating}
\label{sec:variants}
As mentioned above, the fetched information from the memory may repeat some information of the input triples, which causes the content-repeating problem. We investigate several variants of our core model iMAG to solve this problem.

\textit{\textbf{Repeating Sentence Penalty (RSP).}}
 The output token sequence $\{t_i\}$ is calculated by taking the maximum probability of $y_i$, namely $t_i = \arg\max(y_i)$. 
Then, the start and the end of the last Longest Repeating Non-overlapping Substring (LRNS) are calculated by applying the LRNS algorithm on $\{t_i\}$. LRNS is calculated as:
\begin{equation}
\begin{split}
LRN&S(S)\overset{\Delta}{=}\arg\max_{p,q} (q-p)\\
s.t.\ \ \ &1 \le p < q \le m\\
& k < 2p-q\\
& S[p+r]=S[k+r]\ \ \ 0 \le r \le q-p,
\end{split}
\end{equation}
in which $m$ is the length of the output sequence $S$. $p,q\in \mathbb{N}$ are index variables representing the start and end indexes for the LRNS algorithm. It can be solved with dynamic programming with the complexity of $\mathcal{O}(n^2)$.

Then the repeating sentence loss $\mathcal{L}_{rsp}$ is defined as the sum of the probability of each repeating token found by the LRNS algorithm:
\begin{equation}
    \mathcal{L}_{rsp}=\sum_{r=p}^{q} y_r[t_r],
\end{equation} where $y_r[t_r]$ is the probability value of $t_r$ in the distribution $y_r$. Finally, the total loss of this variants is defined as the weighted sum of $\mathcal{L}_{s2s}$ and  $\mathcal{L}_{rsp}$:
\begin{equation}
    \mathcal{L}=\mathcal{L}_{s2s}+\alpha \mathcal{L}_{rsp},
\end{equation}
in which $\alpha$ is a weight coefficient.

\textit{\textbf{Repeating Word Penalty (RWP).}}
Instead of penalizing for the repeated substring as described above, we can also penalize for the repeating words. This is inspired by the observation that in an ordinary sentence, a word may not appear too many times. In a generated sequence, each word will have a probability. We accumulate the probability of each word in the whole sequence and penalize it if the probability sum of a certain word is greater than a threshold $\gamma$. The loss can be expressed as:
\begin{equation}
\mathcal{L}_{rwp}=\mathbf{1}^T\max(\mathbf{0},\sum_{i=1}^m y_i - \gamma\mathbf{1}),
\end{equation}
in which $\mathbf{0}$ and  $\mathbf{1}\in \mathbb{R}^{|V|}$, $\max(\cdot,\cdot)$ returns a vector that takes the element-wise maximum of the two input vectors. 

\textit{\textbf{Reinforcement Learning (RL).}}
RL algorithm can optimize the parameters according to the model performance, no matter whether it is differentiable. We can use the LRNS score as the reward function.
Similar to \cite{yu2017learning,narayan2018ranking}, we apply RL to the generated sequence. Specifically, the REINFORCE algorithm \cite{williams1992simple} is used to calculate the gradient of the parameters based on the probability of each output word and the LRNS length ratio. The gradient can be expressed as:
\begin{equation}
\nabla_{\theta,\psi} \mathcal{L}_{RL}(\theta,\psi)=\mathbb{E}[\nabla_{\theta,\psi} \sum_{i=1}^m \log y_i[t_i] \frac{m-q+p}{m}],
\end{equation}
in which $p$ and $q$ are calculated by the LRNS algorithm and we use the length ratio $\frac{m-q+p}{m}$ as the reward function of the RL algorithm.

\textit{\textbf{Coverage (CVG).}}
We employ the coverage mechanism \cite{tu2016modeling,gehrmann2018end} which is capable of solving the repetition problem in neural machine translation systems. It averages the attention in the decoder input and thus suppresses generating repeating content.

\subsection{Learning and Prediction}
The training of our core model iMAG is straightforward, however, there are two major problems when training its variants.
Firstly, it suffers from the exposure bias problem \cite{ranzato2016sequence,paulus2018a}. In the traditional seq2seq framework, the output is generated in a teacher-forced manner. Precisely, the ground-truth word in the training sequence at each time step, instead of the predicted one, is fed into the next time step. The benefit is that it can avoid error propagation and enable parallel training. However, it hides the content-repeating problem in the training stage. Consequently, it makes the model unable to deal with this problem in testing since it has not learned that capability. Secondly, if we train the model without the teacher-forced manner, the procedure becomes impossible to be parallelized since each step needs the output of the previous step, which makes the training extremely slow.

In order to overcome the above dilemma, we train the model in two separate stages. In the first stage, we train the seq2seq network without the repeating sentence penalty in a teacher-forced way. After certain epochs, we switch the training to using the repeating sentence penalty and removing the teacher forcing.

\section{Experiments}
\subsection{Dataset}
\label{sec:dataset}  

We adapt WebNLG v2 \footnote{ \mbox{https://gitlab.com/shimorina/webnlg-dataset}} and derive a new dataset suitable for the new problem setting. It simply resamples the training/testing partition and can thus be easily extended to other datasets in a similar manner. For the training data, we draw (triples, text) pairs that contain at least two triples to make it possible to drop some triples. 
There is no gold standard target text for these triples since we hope the model can decide the most suitable knowledge to retrieve and present. Therefore, the models will be evaluated by a pseudo target sentence and human evaluation which is discussed in Sec. \ref{sec:exp-setup}.

\begin{table}[t]
  \centering
  
  \scriptsize
\begin{tabular}{@{~}l@{~}|@{~}lll}
\toprule

 {} & Train &    Dev &   Test \\
\hline
 \# (triples, text) pairs  &  33,608 &  706 &  1,501 \\
 Avg. triples per pair &  1.512 &  1.508 &  1.512 \\
 Avg. target length  &  27.23 &  - &  - \\
\bottomrule
\end{tabular}

\caption{Statistics of our dataset. Length is measured by word count.}
\label{tab:dataset}
\end{table}

\begin{table*}[t]
  \centering
\scriptsize
\begin{tabular}{r|l@{~}@{~}l@{~}@{~}l@{~}@{~}l@{~}@{~}l@{~}@{~}l@{~}@{~}l@{~}@{~}l@{~}@{~}l@{~}@{~}l@{~}@{~}l@{~}@{~}l}
\toprule
{} &     R$_L$ &     P$_L$ &     F$_L$ &     R$_2$ &     P$_2$ &     F$_2$ & R$_{SU4}$ &     P$_{SU4}$ &     F$_{SU4}$  & LEN &     LRNSR$\downarrow$ &   DRATE$\uparrow$ \\
\hline
S2S            &  0.234 &  0.538 &  0.317 &   0.164 &  0.399 &  0.225 &   0.147 &  0.407 &  0.206 &   23.7 &  0.084 &  0.807 \\
S2SF           &  0.187 &  0.713 &   0.290 &   0.134 &  0.570 &   0.210 &   0.104 &  0.600 &   0.170 &   13.4 &  0.056 &  0.916 \\
iMAG            &  0.281 &  0.428 &  0.335 &   0.192 &  0.299 &  0.230 &   0.186 &  0.301 &  0.226 &   35.4 &    0.120 &  0.670 \\
\hline
iMAG+CVG      &  0.288 &   0.420 &  0.336 &   0.195 &  0.289 &  0.229 &   0.189 &  0.292 &  0.225 &   37.8 &   0.129 &  0.649 \\
iMAG+RL       &  0.269 &  0.436 &  0.327 &   0.182 &  0.301 &  0.223 &   0.177 &  0.309 &  0.220 &   33.3 &   0.111 &  0.693 \\
iMAG+RWP      &  0.276 &   0.430 &  0.331 &   0.187 &  0.298 &  0.226 &   0.181 &  0.302 &  0.221 &   34.5 &   0.112 &  0.690 \\
iMAG+RSP      &  0.271 &  0.441 &  0.330 &   0.185 &  0.309 &  0.227 &   0.178 &  0.314 &  0.222 &   33.5 &   0.102 &  0.706 \\
\hline
iMAG w/o memory &  0.135 &  0.486 &  0.206 &  0.073 &  0.287 &  0.114 &  0.065 &  0.334 &  0.105 &   14.5 &  0.063 &  0.886 \\
iMAG w/o source &  0.168 &  0.452 &  0.240 &  0.075 &   0.210 &  0.108 &  0.078 &  0.262 &  0.116 &   21.9 &    0.110 &  0.729 \\
\bottomrule
\end{tabular}

\caption{Main results on our ITG dataset. $\downarrow$: smaller is better, $\uparrow$: larger is better.}
\label{tab:result}
\end{table*}

\subsection{Comparison Models and Our Variants}
We compare our model with two baselines. 
\textbf{S2S} uses vanilla seq2seq \cite{sutskever2014sequence, cho2014learning} framework with the standard attention and copy mechanisms.
\textbf{S2SF} is the same as S2S, but utilizes the original WebNLG without removing any triples for training. For testing and development, it uses the same data as other methods. 
We also examine different variants of our model.
\textbf{iMAG} is the core model of our framework that contains the information memory without any component dealing with the content-repeating problem, while
\textbf{iMAG+CVG}, \textbf{iMAG+RL}, \textbf{iMAG+RWP}, and \textbf{iMAG+RSP} are variants that employ the methods presented in Sec. \ref{sec:variants} to deal with this problem.
We also conduct two ablation experiments.
\textbf{iMAG w/o memory} employs our iMAG model at the training phase, but for testing, the information memory is deliberately omitted. Therefore, this model allows us to examine how much information can be generated only with the input triples.
\textbf{iMAG w/o source} also has the same training phase as iMAG, but for testing, its seq2seq network excludes the input triples and only uses the information memory (of course, queried with the input triples). Therefore, this model allows us to examine whether the memory can give correct information about the input entities.

\subsection{Experimental Setup}
\label{sec:exp-setup}
Evaluating the generated text is challenging since there are no gold standard target sentences. We propose to evaluate the models by comparing with the pseudo target sentence, reporting several automatic evaluation metrics, and conducting a human evaluation. The pseudo target sentence is made from those 1-triple data samples in the original WebNLG dataset. Each data sample contains one triple and one corresponding sentence and all of them have not been included in the training set. The pseudo target sentence is made by concatenating all 1-triple sentences describing any of the entities in the given input triple set. For example, if the entity ``\textit{Bill Gates}'' and ``\textit{Microsoft}'' are included in the input triple set, the pseudo target sentence will concatenate all 1-triple sentences describing them as ``\textit{Bill Gates was born in Seattle. Bill Gates is a US citizen. Microsoft was founded in Seattle. ...}''. Several metrics are employed for the comparison, namely ROUGE$_L$, ROUGE$_2$, ROUGE$_{SU4}$, LEN, LRNSR, and DRATE. In the ROUGE metric \cite{lin2004rouge}, P, R and F (precision, recall and F-value) are all reported.
LEN denotes the average length of the output. LRNSR stands for LRNS Ratio and shows how much duplicated information is included in each sentence. It is defined as the length of the LRNS divided by the length of the sentence. DRATE stands for distinctness rate which is defined as the ratio of the number of distinct words over the sentence length. The smaller the metric is, the less information is contained in the sentence.
Note that instead of BLEU, we report ROUGE-P as the precision-oriented metric because the reference is too long and the brevity penalty is close to 0 making the BLEU scores close to 0. 
We keep all the hyperparameters in \citet{opennmt} unchanged. We use grid search to tune hyperparameters on the development set. We choose CNN kernel size $k=3$ from $\{2, 3, 5, 7\}$, memory size $l=600$ from $\{200, 400, 600, 800, 1000\}$, $\alpha=0.5$ from $\{0.2, 0.5, 1.0, 2.0\}$, and $\zeta=0.5$ from $\{0.1, 0.2, 0.5, 0.7, 0.9\}$. We iterate 20K batches (50 samples each batch) with teacher forcing mechanism and then remove such mechanism and train for another 40K batches which is choosed from $\{10K/50K, 20K/40K, 30K/30K, 40K/20K, $ $50K/10K\}$. It takes 3.8 hours to train the model.

\subsection{Experimental Results}

The experimental results are shown in Table \ref{tab:result}. It can be concluded from the results that our proposed models can generate valid sentences containing informative knowledge. The F values show that in general, the iMAG model can generate more favorable text which has a better balance between precision and recall. 
Considering the R values and the LEN values together, we can conclude that the advantage in F values attributes to the contribution of our information memory. It indeed memorizes useful knowledge in training and helps generate longer and meaningful text at the cost of affecting some precision. Another cause of the relatively lower precision is that the pseudo target sentence is concatenated by 1-triple sentences which may be different from the generated style.  
On the other hand, we find that S2SF achieves the best precision but with the smallest LEN. The reason is that S2SF is trained with strictly aligned (triples, text) pairs and it will not output additional information about the entities during testing, thus it has less chance to get wrong. 
It can also be observed that the S2S and iMAG model generate longer sentences than S2SF. This is because, in the training phase, S2S and iMAG models are trained to retrieve related information from the memory while S2SF model does not learn such capability. As a result, the S2SF model can only generate text exactly describing the input triple set when testing.

\begin{table}[t]
  \centering
  \scriptsize
  
  \begin{tabular}{l|lll}
  \toprule
  {} & Plausibility & Grammaticality & Richness \\
  \hline
  S2S        &         6.05 &           6.01 &     5.84 \\
  S2SF       &         6.11 &           6.05 &     5.32 \\
  iMAG + RSP &         5.84 &           6.05 &     7.02 \\
  \bottomrule
  \end{tabular}

\caption{Human Evaluation.}
\label{tab:human-eval}
\end{table}

LRNSR and DRATE are two metrics that can more intuitively illustrate the characteristics of the generated text. While iMAG outputs longer and informative text, it also produces more repetitions. 
In the second part of Table \ref{tab:result}, we examine the effect of those four variants for alleviating the content-repeating problem. 
The results of iMAG+CVG show that the coverage mechanism performs poorly. Precisely, both LRNSR and DRATE become worse. 
This is because the fetched memory information may have the same information as the input triples, averaging attention to each hidden vector of the encoder cannot help solve the content-repeating problem caused by the repetitions in the encoder input sequence.
The results of iMAG+RL show that the reinforcement learning (RL) module here can help solve the problem. However, the improvement is not significant because the RL module depends on the sampling of the output sequences. If the probability is low, the gradient is also very small and thus it is hard to converge.

iMAG+RWP and iMAG+RSP, which are the two variants directly penalizing the repetitions, are more effective in solving the problem. iMAG+RSP performs better, and its LRNSR value drops to 0.102 while DRATE increases to 0.706. 
iMAG+RWP penalizes nearly all repeating words, and thus it is sometimes too harsh since high-frequency words can appear multiple times and should not be penalized.
iMAG+RSP, penalizing on the longest repeating common substring, is a more favorable solution.

The third part of Table \ref{tab:result} gives the results of the two ablations.
It shows that the trained integrated model will not perform well if one type of the input is removed. 
The results of ``iMAG w/o memory'' show that without the information memory, the model can only generate a short sentence. Meanwhile, its LRNSR measure drops dramatically to 0.063 and the DRATE increases to 0.886, which confirms that the repeating is mainly caused by the fact that the fetched memory repeats some information of the input triples.

\begin{table}[t]
  \centering
  \scriptsize
  
  \begin{tabular}{l|llll}
  \toprule
  {} & CR & PC & CC & AR \\
  \hline
  S2S        &        0.819 &     0.658 &    0.879 &      1.38 \\
  S2SF       &        0.838 &     0.528 &    0.852 &     0.73 \\
  iMAG + RSP &        0.803 &     0.676 &    0.889 &      1.78 \\
  \bottomrule
  \end{tabular}

\caption{Acquisition analysis.}
\label{tab:acquisition}
\end{table}

\begin{table}[t]
  \centering
  
  \scriptsize
\begin{tabular}{@{}r@{~}|@{~}l@{~}@{~}l@{~}@{~}l@{~}@{~}l@{~}@{~}l@{}}
\toprule
{} & R$_L$ & P$_L$ & F$_L$ & LEN &     LRNSR  \\
\hline
S2S       &  0.148 &  0.479 &  0.223 &  15.2 &   0.074 \\
S2SF      &  0.117 &  0.757 &  0.198 &  7.68 &   0.119 \\
iMAG       &  0.183 &  0.478 &  0.261 &  21.2 &  0.079 \\
iMAG+RSP &  0.183 &  0.479 &   0.260 &    21.0 &  0.077 \\
\bottomrule
\end{tabular}

\caption{Results of entity biography generation.}
\label{tab:entity}
\end{table}

\begin{table*}[t]
  \centering
  
  \scriptsize
\begin{tabular}{@{~}l@{~}|@{~}L{13.6cm}@{~}}
\toprule
Input Entity & Sentence \\
\hline
Allen Forrest & Allen Forrest is a solo singer who plays acoustic music. He was born in Fort Campbell, KY and was raised in Dothan, AL .\\
\hline
109 Felicitas & 109 Felicitas, which has an epoch date of December 31st 2006, has a periapsis of 283326000000.0 and an orbital period of 139705000.0 .\\
\hline
Alhambra & The Alhambra has a ship beam of 8.3m and is 63800.0 millimetres long. It was launched on 31st May 1855.\\
\midrule
\end{tabular}

\caption{Case study for entity biography generation. }
\label{tab:biography}
\end{table*}

\subsection{Human Evaluation}

To further evaluate the generation quality, we conduct a human evaluation focusing on plausibility, grammaticality, and richness. Specifically, plausibility refers to the correctness of the generated sentence. The grammaticality refers to the grammatical correctness while the richness refers to how much new correct information is added. Human raters are asked to score the generated sentences from 1 to 10 in three different perspectives.
The results are shown in Table~\ref{tab:human-eval}. It can be concluded that the iMAG+RSP outperforms S2S and S2SF models significantly in richness showing that our proposed model is capable of generating more informative text. 
The plausibility of the iMAG+RSP model falls a little behind other models since generating more content makes it easier to make mistakes. However, quantitatively, the hallucination generation problem \cite{tian2019sticking} is not much severe than other models showing the model can retrieve correct knowledge. 
The grammaticality scores are almost the same for all models showing that all of these models can generate grammatically correct sentences.

\begin{table*}[!t]
  \centering
 
  \scriptsize
\begin{tabular}{R{1.3cm}|L{13cm}}
\toprule
Model & Sentence \\
\hline
S2S            &  Aaron Turner played with the bands Twilight \textcolor{cyan}{and Old Man Gloom} . \\
\hline
S2SF           &  Aaron Turner played with the band Twilight . \\
\hline
iMAG            &  Aaron Turner was \textcolor{cyan}{born in the United States and started performing in 1995 }. He played with the band Twilight and is \textcolor{cyan}{associated with the group Greymachine} . \textcolor{orange}{He played with the band Twilight }. \\
\hline
iMAG+ RSP      &  Aaron Turner is \textcolor{cyan}{an artist for the band Isis} and \textcolor{cyan}{is associated with the group Greymachine} . He played with the band Twilight , \textcolor{cyan}{the musical genre of which , is black metal }. \\
\bottomrule
\end{tabular}

\caption{Case study for the input ``$<$Aaron Turner, associatedBand, Twilight ( band ) $>$''. \textcolor{cyan}{Blue} font: information not mentioned in the input triple. \textcolor{orange}{Orange} font: repeating sentence. 
}
\label{tab:casestudy}
\end{table*}

\subsection{Acquisition Analysis}
In order to quantitatively evaluate  how much knowledge and which knowledge can be obtained from the memory, we conduct an acquisition analysis as shown in Table \ref{tab:acquisition}. Given a generated sentence corresponding to an input triple set, we say it covers a 1-triple data sample if it contains 80\% of the words in the 1-triple sentence. We analyze the relevance between the input triple's relations and the relations covered by the generated text. We denote $r_j^{i}$ as the $i$th relation in the $j$th input triple set and $\hat{r}_j^{k}$ as the $k$th covered relation for the $j$th generated sentence. The co-occurrence ratio (CR) is defined as the portion of the $(r_j^{i}, \hat{r}_j^{k})$ pairs that co-exist in any of the input triple set of the training set. The pair correlation (PC) is defined as the correlation coefficient between the count of the pair $(r_j^{i}, \hat{r}_j^{k})$ and the count of its co-occurrence in the training input triple set. The count correlation (CC) is defined as the correlation coefficient between the count of the relation $\hat{r}_j^{k}$ and the count of it in the training input triples. All of the CR, PC, CC metrics give a quantitative analysis of which relation is preferred when retrieving from the memory. It can be concluded from the results that if two relations are closely connected in the training set, the model is more likely to retrieve one relation from the memory if the other is given. The acquisition ratio (AR) is defined as the count of the covered new relations which does not contain the relations in the input triple set, over the count of the input triples. It gives an intuition of how much knowledge on average the model will retrieve from the memory. From the result, we can observe that the iMAG+RSP model retrieves more relations than other models.

\subsection{Entity Biography Generation}

To show that our model is capable of storing useful information into the memory, we experiment on generating an entity's biography by only inputting the entity name. The numerical results are shown in Table \ref{tab:entity} while some generated example biographies are shown in Table \ref{tab:biography}. 
Both iMAG and iMAG+RSP can generate longer descriptions than that generated by S2S and S2SF. The reason is the same as above. Concretely, S2S and S2SF heavily depend on the information from the input triples while our model is capable of utilizing the information from the memory. Moreover, our proposed model can generate real background knowledge about the given entities instead of hallucination generation \cite{tian2019sticking}.

Compared with Table \ref{tab:result}, the P$_L$ values of iMAG and iMAG+RSP in Table \ref{tab:entity} are higher. The reason is that for this experiment, we do not have the relation name as part of the input which may fetch some information about the relation of other entities, while the reference here is only about the input entity.
In Table \ref{tab:entity}, all the metrics for iMAG and iMAG+RSP are very similar. The reason is that we only have the entity name as input and thus the information redundancy in the input sequence (i.e. containing similar information with the fetched memory) of the seq2seq network does not exist anymore. 
Table \ref{tab:biography} shows the generated biographies of three entities. We can see that each biography contains rich information about the entity. In the original data, such information exists as multiple small sentences and here they are reasonably merged into longer and well-formatted sentences.

\subsection{Case Study of Generated Text for Triples}

In order to provide a more intuitive understanding of how our models are capable of generating sentences with extra useful information, we analyze the output result of the input triple
``$<$Aaron Turner, associatedBand, Twilight ( band ) $>$''. The triple only gives the basic information of the artist and the models are trying to give more background knowledge about Aaron Turner and Twilight. Table \ref{tab:casestudy} shows the generations of different models.

It can be observed from Table \ref{tab:casestudy} that instead of hallucination generation \cite{tian2019sticking}, our proposed models iMAG and iMAG+RSP can successfully add useful extra information about the entities in the given triple. They infer from the memory that Aaron Turner is in Isis and associated with the group Greymachine, the musical genre of Twilight is black metal, etc. 
On the other hand, the S2S model can also output some extra information that was encoded in the network. However, due to the deprivation of using auxiliary modules to store the information, S2S fails to provide extra messages about Aaron Turner.
The S2SF model fails to introduce extra information for the entities, because it has neither been trained to add information nor does it have any extra component to store the extra information.
We also observe that the iMAG model suffers from the content-repeating problem, while with the help of RSP, iMAG+RSP can learn to overcome the problem.

\section{Conclusions and Future Work}
We studied a new problem setting of KB-to-text, namely informative text generation, which requires models to generate text that not only conveys the input triples but also depicts the extra information of the involved entities. 
To solve the problem, we propose a novel memory augmented generator which has a specifically designed information memory to store and utilize the extra information. 
The experimental results show that our model can generate informative sentences from the input triples by fetching extra related information that is not mentioned by the input triples.
Moreover, our model is capable of generating fluent biographies by only using the memory, and the proposed variants can effectively alleviate the content-repeating problem.

The following directions remain to be explored: (1) Our model only needs to make minor changes to the existing dataset and can thus be naturally extended to other datasets. (2) We only focus on resolving the task in specific domains and the open domain setting can also be explored in the future.

\bibliography{reference.bib}
\bibliographystyle{acl_natbib}

\appendix

\end{document}